\documentclass{article}

\usepackage[english]{babel}

\usepackage[letterpaper,top=2cm,bottom=2cm,left=3cm,right=3cm,marginparwidth=1.75cm]{geometry}

\usepackage{amsmath}
\usepackage{graphicx}
\usepackage[colorlinks=true, allcolors=blue]{hyperref}
\usepackage[noend]{algpseudocode}
\usepackage{algorithmicx,algorithm}
\usepackage{amssymb}
\usepackage{authblk}
\title{Faster and Simpler Siamese  Network for Single Object Tracking}

\author[a,b]{Shaokui Jiang}
\author[a,b]{Baile Xu}
\author[c]{Jian Zhao}
\author[a,b]{Furao Shen \thanks{Corresponding author: frshen@nju.edu.cn}}

\affil[a]{State Key Laboratory for Novel Software Technology,
Nanjing University,
163 Xianlin Avenue, Qixia District,
Nanjing, 210023, China}
\affil[b]{Department of Computer Science and Technology,
Nanjing University,
163 Xianlin Avenue, Qixia District,
Nanjing, 210023, China}
\affil[c]{ School of Electronic Science and Engineering,
Nanjing University,
163 Xianlin Avenue, Qixia District,
Nanjing, 210023, China
\authorcr Email: jiangshaokui@forxmail.com, \{blx, jianzhao, frshen\}@nju.edu.cn}
\date{}
\begin{document}
\maketitle

\begin{abstract}
 Single object tracking (SOT) is currently one of the most important tasks in computer vision. With the development of the deep network and the release for a series of large scale datasets for single object tracking, siamese networks have been proposed and perform better than most of the traditional methods. However, recent siamese networks get deeper and slower to obtain better performance. Most of these methods could only meet the needs of real-time object tracking in ideal environments. In order to achieve a better balance between efficiency and accuracy, we propose a simpler siamese network for single object tracking, which runs fast in poor hardware configurations while remaining an excellent accuracy. We use a more efficient regression method to compute the location of the tracked object in a shorter time without losing much precision. For improving the accuracy and speeding up the training progress, we introduce the Squeeze-and-excitation (SE) network into the feature extractor. In this paper, we compare the proposed method with some state-of-the-art trackers and analysis their performances. Using our method, a siamese network could be trained with shorter time and less data. The fast processing speed enables combining object tracking with object detection or other tasks in real time.
\end{abstract}

\section{Introduction}

Object tracking is one of the main applications of computer vision. Single object tracking is fundamental to tracking tasks, which has a totally different methodology with multiple object tracking. Object tracking has been widely used in video surveillance\cite{joshi2012survey}, virtual reality\cite{dempski2006arbitrary}, human-machine interaction, image understanding, autonomous driving\cite{benenson2008towards} and so on. It also plays an important role in other related missions. For example, object detection is a relatively more complex application in computer vision, where even state-of-the-art could miss some objects in specific scenes. In that case, object tracking could help. Some missing targets caused by occlusion or light changing could be detected by the assistance of tracking. Besides, detecting objects in each frame of a video is time consuming and unnecessary, whereas tracking detected object could save a lot of time. However, recent tracking algorithms tend to exchange speed for accuracy, which deviates from the original intention of target tracking.

In this paper, we focus on designing an accurate single object tracker that runs fast under the condition of limited hardware resources. Firstly, we adds the Squeeze-and-excitation (SE)\cite{hu2018squeeze} block into the feature extractor as a subnetwork called SE Layer. The SE Layer enables the network to train a suitable number of parameters faster and more precisely. Nextly, the feature maps of video frames are convolved by the kernels consists of the features of the tracked target. Different from other siamese networks\cite{bertinetto2016fully}, the convolution result of our work is smaller. By this means the number of parameters in this layer is reduced, and hence the training and inference processes are greatly accelerated. Finally, we get the output by putting the feature map into our regression part, which consists of a kernel size of $1\times1$ and a fully connected layer. With the output and the size of the search area, we locate the tracked target.

We train the network using only two datasets, GOT\cite{huang2019got} and ImageNet-VID\cite{russakovsky2015imagenet}, which consists of no more than 15000 video clips in total. The network is well trained within 5 training epochs. We evaluate our method on VOT2015\cite{kristan2015visual}, VOT2016\cite{vot2016}, VOT2017\cite{kristan2017visual} and OTB100\cite{WuLimYang13} challenges. The results show great performance and excellent running speed on low-end hardware.
\begin{figure*}[ht]

    \centering
    \includegraphics[scale=0.4]{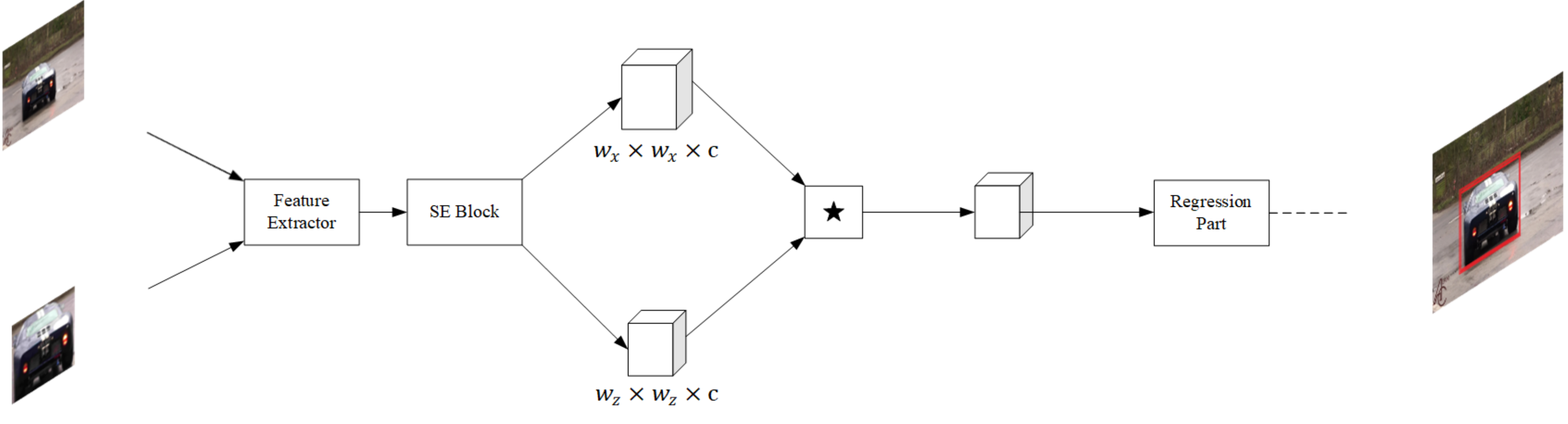}
    \caption{This is the framework of our network. The feature extractor takes RGB images as input in both branches with shared parameters. After extracting, the feature maps are fed into the SE block. The SE block adjusts the features extracted in different channels. The template features after extracting and weighting are $w_z\times w_z\times c$ in shape, the detection features from the same way are $w_x\times w_x\times c$. $\star$ denotes correlation operator, which works for every channel and generates a feature map of $(w_x-w_z+1)\times(w_x-w_z+1)$ with $c$ channels. Finally, the regression part turns the feature map to the position of the target in the detection frame. The output is the relative position for the target exactly.}
    \label{fig1}
\end{figure*}

\section{Related Work}
\label{sec:related}
Visual object tracking can be traced back to the 1980s. Most traditional algorithms were proposed around the year of 2000. Comaniciu and Meer\cite{comaniciu2003kernel} used the mean-shift procedure searching for the target by comparing histograms of the target and the search area. Beymer and Konolige\cite{beymer1999real} introduced Kalman filtering to the object tracking for predicting the position of the target. Particle filters were used in multiple object tracking in 2002 by Hue\emph{et al}\cite{hue2002tracking}. TLD (Tracking-learning-detection) proposed by Kalal\cite{kalal2011tracking} is a typical algorithm that unites object detection with object tracking, which works well in long-term object tracking. TLD could learn the model and parameters in the progress of tracking and detection, which is controversial because most of the trackers only use the detection in the first frame.

MOSSE (Minimum Output sum of Squared Error filter) proposed by Bolme\cite{bolme2010visual} is the first work that introduces correlation filter to object tracking. MOSSE started the next 10 years research boom for correlation filter in this task. Inspired by MOSSE, in the year of 2015, Henriques \emph{et al}.\cite{henriques2014high} put forward KCF (Kernelized Correlation Filters) algorithm. KCF replaces the sliding windows with cyclic shifts, which enables the calculation be done in Fourier space. KCF speeds up the algorithm greatly. In addition, KCF uses ridge regression as the loss function. The ridge regression in linear spaces is simplified by projecting the calculation to non-linear spaces using kernel functions. More works about kernelized correlation filters, such as Martin\cite{danelljan2014accurate}, used scaling pool to solve the problem of target shift because of the changing size in KCF. Liu\cite{liu2015real} proposed part-based visual tracking via adaptive correlation filters.

After Alexnet\cite{krizhevsky2012imagenet} was designed in 2012, deep learning was widely used in various fields, including object tracking. GOTURN (Generic Object Tracking Using Regression
Networks)\cite{held2016learning} proposed by Held adopts siamese network for feature extraction. After the feature extraction, GOTURN contacts the features directly for the regression in next steps. GOTURN runs fast because of the simple architecture. SiameseFC\cite{bertinetto2016fully} proposed by Bertinetto \emph{et al}. is similar to the GOTURN in the structure of feature extractor, but SiameseFC put forward to convolving template features on the features of search area, instead of stacking them together. SiameseFC achieves state-of-the-art performance in multiple benchmarks, which exploits the power of deep convolutional networks for object tracking. After SiameseFC, there are a lot of works in object tracking with correlation filtering and siamese network. 

SiameseRPN\cite{li2018high} points that SiameseFC do not regress the position of the tracked target totally. Instead, SiameseFC calculates the result using the rough position from heat map and multiple shape. This is one of the reasons why the accuracy of SiameseFC was not high enough. SiameseRPN makes a better use of the siamese network in object tracking, introducing the region proposal network into object tracking for replacing the part after feature extractor. SiameseRPN and other related papers, DaSiameseRPN\cite{zhu2018distractor} and SiameseRPN++\cite{li2019siamrpn++}, all achieve excellent performance. SiamMask\cite{wang2019fast} combines object tracking and segmentation into one network but two branches. SiamMask uses deeper feature extraction network, which runs slowly but gets better performance. 


The attention mechanism in deep learning has been widely used in RNN and CNN networks. The visual attention model is trying to let the neural network be able to “focus” its “attention” on the interesting part of the image where it can get most of the information, while paying less “attention” elsewhere. Non-local networks\cite{wang2018non} presents non-local operations for capturing long-range image features. Non-local operation means computing the response of a point as a function of the features of all points. The Squeeze and Excitation (SE)\cite{hu2018squeeze} block can be stacked together with any multi-channel layers, assigning different channels with different weights. The SE block brings significant improvements for existing state-of-the-art CNNs. Another work extends the layer weights to the channel space and the spatial space\cite{woo2018cbam}, but the main idea is almost consistent with the SE block.

\section{Proposed Approach}
\label{sec:proposed}
\subsection{Method Overview}

In this section, we explain our approach in detail. The framework of our model is illustrated in Fig. \ref{fig1}. The framework consists of four parts: 
\begin{enumerate}
\item Feature extractor: combinations of convolution layers, pooling layers and batch normalization layers;
\item SE layer: giving different channels their corresponding weights;
\item Convolution operation: a convolution operation simulates the progress of searching the tracked object in a video frame;
\item Regression part: a combination of a convolution kernel and a fully connected layer turns the result comes from the last step into a vector that outputs the target region.
\end{enumerate}

We feed the template patch and the detection patch from each frame of the video into the network as shown in Fig. \ref{fig1}. The network outputs the relative position of the target in the detection patch. Our training procedure is offline, which means there is no time spent on model training during the testing phase. 

\subsection{Feature Extractor}
The feature extractor has two branches with shared parameters: the template branch takes the image patch containing the target as input, which is denoted as $z$; the detection branch takes the patch $x$ from a video frame as input. The layers in the feature extractor are adopted in both the template branch and the detection branch with shared parameters. These layers aim to extract the features that make greatest response in two similar patches, regardless of which branch it is belong to. We denote $F$ as the operation of feature extraction, $F(z)$ as the output of the template branch, and $F(x)$ as the output of the detection branch.

The feature extractor of our work is alternative. It can be replaced by most of the backbones commonly used, such as AlexNet, Resnet\cite{he2016deep} and so on. One thing we should notice is that the complexity of the backbone. According to our experiments, the more complex the backbone, the better the result for tracking. We found that Resnet18 has a better balance between the running speed and the accuracy. Therefore, we recommend you use Resnet18 for the feature extractor.

\subsection{SE Block}
In this section, we introduce the Squeeze-and-Excitation (SE) block inserted in our framework. Inspired by the SE network, we add the block after the feature extractor. The SE block is one of the attention mechanism enabling feature recalibration. With SE block, our model performs better and converges faster. 

\begin{figure}[ht]
\centering
\includegraphics[scale=0.3]{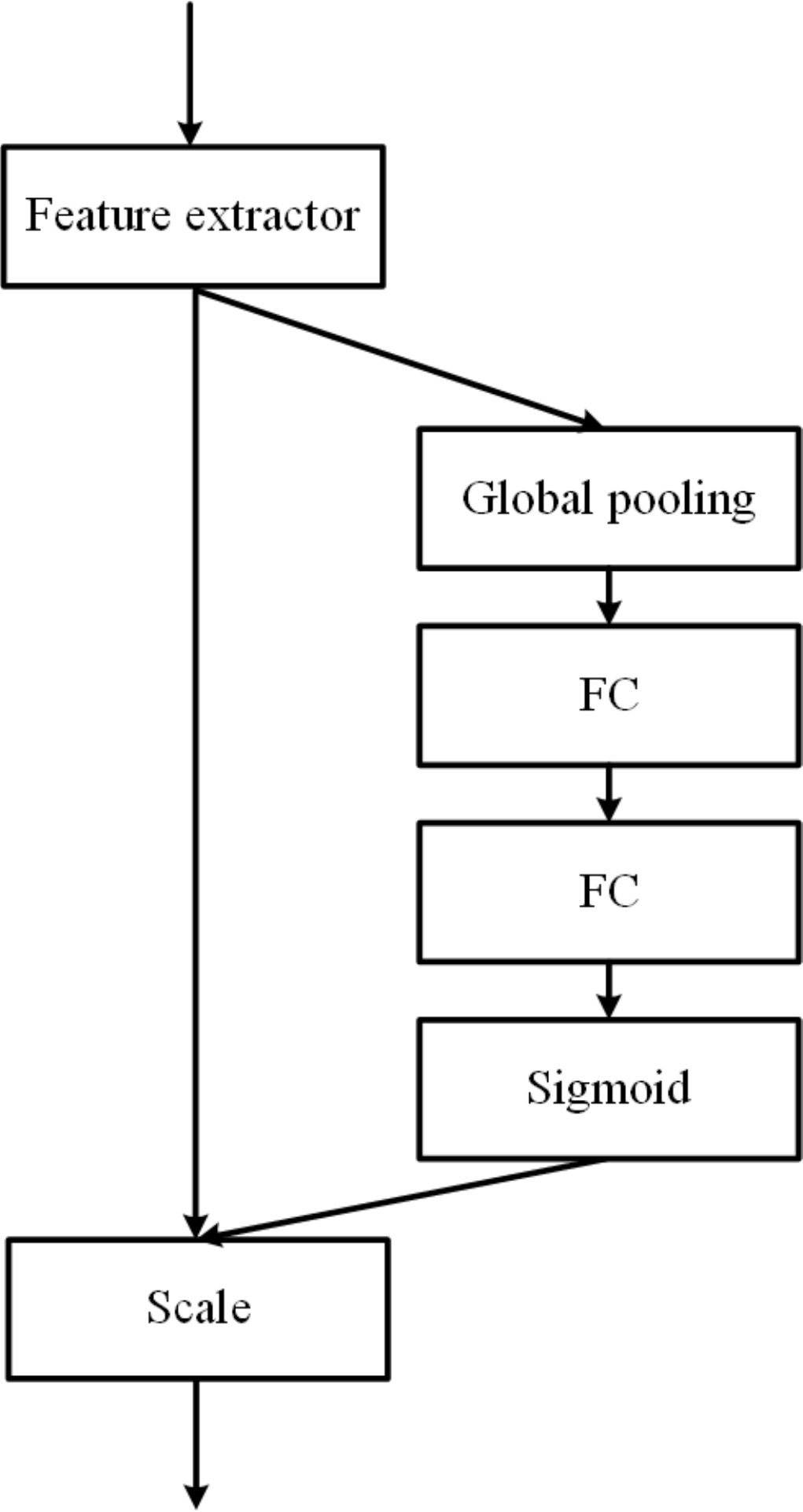}
\caption{The structure of the SE block, which contains the vector of the same length as the number of channels. We acquire this vector by training the sequences of activation functions and fully connected layers. The final feature map is obtained by multiplying the vector to relevant channels.}
\label{fig2}
\end{figure}

The SE block has a simple structure, which consists of an average pool, fully connected layers and activation functions as shown in Fig. \ref{fig2}. The input of the SE block is the feature maps come out of the feature extractor, and its output keeps the same size with the input.
We denote the above transformation as $S$, the output for both branches could be expressed as:
\begin{equation}
\label{con:eq1}
\left\{
\begin{array}{lr}
     &F_x=S(F(x)) \\
     &F_z=S(F(z))
\end{array}
\right.
\end{equation}

\noindent where $F_x$ denotes the output of the SE block in detection branch, while $F_z$ denotes the one in template branch.

\subsection{Convolution operation}
In this section, we introduce the convolution operation between the results of two branches. As mentioned earlier, the searching feature map is convolved by the kernel consists of the template feature. We make the convolution at every channel separately. The convolution process of the c-th channel is depicted as Eq. \eqref{con:eq2}. 
\begin{equation}
    Conv_c(F_x,F_z )=F_x^c * F_z^c \label{con:eq2}
\end{equation}
  \noindent where $F_x^c$ denotes the $c$-th channel for $F_x$, while  $F_z^c$ denotes the $c$-th channel for $F_z$. * denotes correlation operator.

The features from detection branch $F(x)$ after extracting and weighting is $w_x\times w_x\times c$ in shape, the template features $F(z)$ from the same way is $w_z\times w_z\times c$. The convolution works for every channel, the result after convolution is $(w_x-w_z+1)\times(w_x-w_z+1) \times c$, which contains the information of the target position in detection patch. The relative position based on the feature map is obtained by feeding the convolution result into the regression part. 

\subsection{Regression part}
In this section, we introduce the regression part in our framework. As we mentioned before, the regression part generates the relative position based on the feature map. This is the most important part in the tracking model, especially for the siamese network. The regression part turns the feature map after convolution to the final position of the target.
Different siamese networks have different implementations of the regression part. For example, SiamFC makes a heatmap using convolution result to locate the target. SiamRPN moves the RPN network in Faster-RCNN\cite{ren2016faster} to regress the x, y, w, h and the probabilities of foreground and background by generating lots of anchors.

\begin{figure}[ht]
\centering
\includegraphics[scale=0.5]{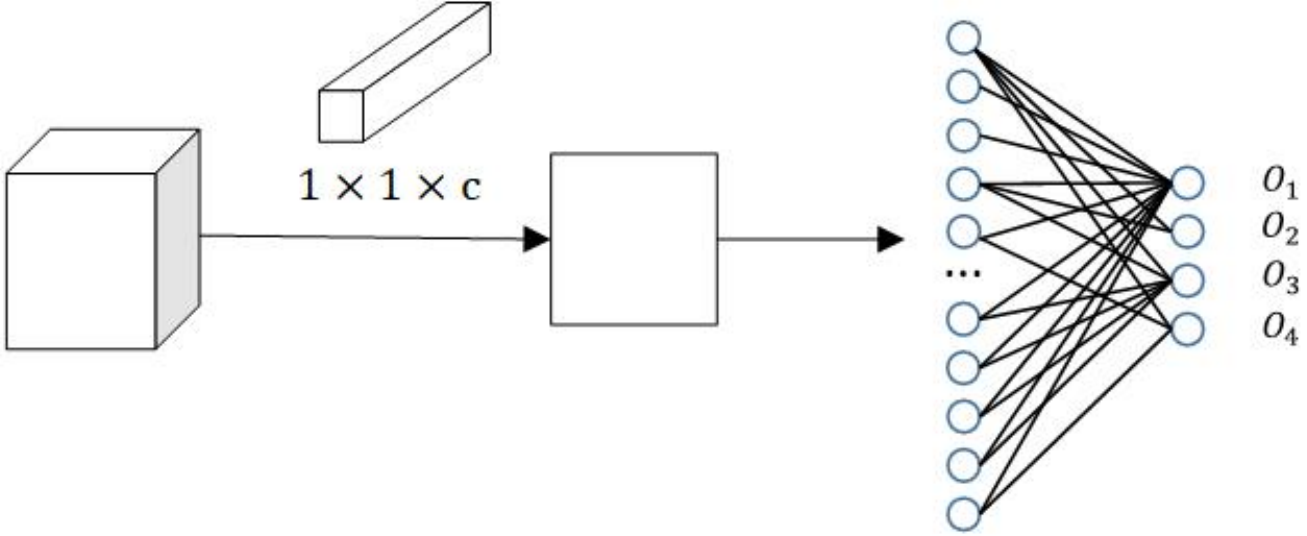}
\caption{Structure of our regression part}
\label{fig3}
\end{figure}

In our network, we propose a brand new approach for regression, just as shown in Fig. \ref{fig3}, which is lightweight and quite simple. We use a convolution kernel of size $1\times1$ to train the weight for different channels. After this step, the shape of the result is $(w_x-w_z+1)\times(w_x-w_z+1)$. Then, we use a fully-connected sub-network to connect the convolution result and the final position. Instead of regressing position directly, we use relative position just like most networks do. The relative position is a vector containing four float numbers, which is denoted as $[O_1,O_2,O_3,O_4]$. The position of target in the detection patch is calculated by Eq. \eqref{con:eq4}:
\begin{equation}
    \label{con:eq4}
    \left\{
    \begin{array}{cc}
         &x_1=O_1*w_f+w_f/2  \\
         &y_1=O_2*h_f+h_f/2  \\
         &x_2=O_3*w_f+w_f/2  \\
         &y_2=O_4*h_f+h_f/2  \\
    \end{array}
    \right.
\end{equation}

\noindent where $x_1, y_1, x_2, y_2$ denote the x-axis coordinate, y-axis coordinate of the top-left corner and the bottom-right corner of the target bounding box in the detection frame. $h_f$ and $w_f$ denote the height and width of the detection patch, respectively.

\section{Training and Inferring}
\subsection{Training Phrase}
In the training phrase, we select two datasets, GOT and ImageNet-VID, for generating samples of the detection frame and the template frame. GOT aims to provide a dataset with wide coverage of common objects for training visual object trackers. GOT covers 560 classes of moving objects and 87 motion patterns. There are more than 10000 video segments provided by GOT with 1.5 million labeled bounding boxes. Another dataset we use is ImageNet-VID, which provides annotations with 30 basic-level categories. 

The datasets mentioned above provide plenty of image sequences. For each sequence, there are annotations of object positions in each image. We process every sequence and corresponding annotation using the five steps as shown in Fig. \ref{fig_select}:
\begin{figure}[ht]
\centering
\includegraphics[scale=0.35]{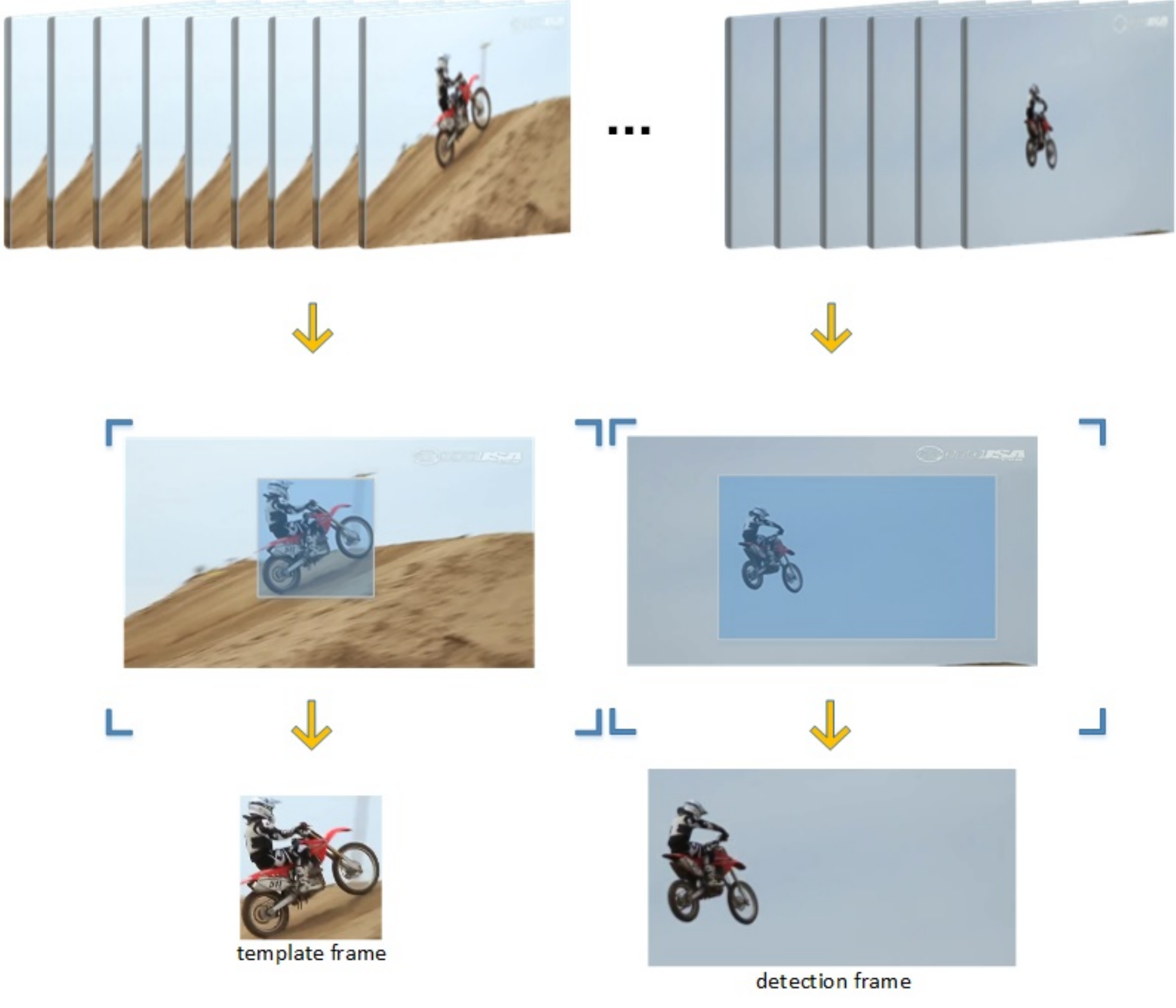}
\caption{Illustration of steps for generating template patches and detection patches from sequences of datasets.}
\label{fig_select}
\end{figure}
\begin{enumerate}
\item Select two frames with a interval randomly selected from 1 to 100. If the target is partly or entirely out of view in either frame, redo this step;
\item Crop the target in one frame as the template patch;
\item For another frame, crop the image as the detection patch according to $left, top, right, bottom$. These values are obstained randomly between the bounding box and the given edge $\hat{left}, \hat{top}, \hat{right}, \hat{bottom}$, just as Eq. \eqref{edge} shown.
\begin{equation}
  \label{edge}
    \left\{
    \begin{array}{cc}
         &left = random(\hat{left}, x_b)  \\
         &top = random(\hat{top}, y_b)  \\
         &right = random(x_2,\hat{right})  \\
         &bottom = random(y_2,\hat{bottom})  
    \end{array}
    \right.
\end{equation}
As for the edge, it is selected randomly from the Eq. \eqref{edge1}, Eq. \eqref{edge2} and Eq. \eqref{edge3} every time;
\begin{equation}
  \label{edge1}
  \left\{
  \begin{array}{cc}
    &\hat{left} = max(0, x_1 $–$ w_b)  \\
    &\hat{right} = min(x_2+w_b,w_f)  \\
    &\hat{top} = max(0, y_1-h_b)  \\
    &\hat{bottom} = min(y_2+h_b,h_f)  \\
  \end{array}
  \right.
\end{equation}

\begin{equation}
  \label{edge2}
  \left\{
  \begin{array}{cc}
    &\hat{left} = max(0, x_1$–$w_b/2)  \\
    &\hat{right} = min(x_2+w_b/2,w_f)  \\
    &\hat{top} = max(0, y_1-h_b)  \\
    &\hat{bottom} = min(y_2+h_b,h_f)  \\
  \end{array}
  \right.
\end{equation}

\begin{equation}
  \label{edge3}
  \left\{
  \begin{array}{cc}
    &\hat{left} = max(0, x_1$–$w_b)  \\
    &\hat{right} = min(x_2+w_b,w_f)  \\
    &\hat{top} = max(0, y_1-h_b/2)  \\
    &\hat{bottom} = min(y_2+h_b/2,h_f)  \\
  \end{array}
  \right.
\end{equation}

\noindent $x_1, y_1, x_2, y_2$ denote the x-axis coordinate, y-axis coordinate of the top-left corner and the bottom-right corner of the target bounding box in current frame respectively. $w_f$ denotes the width of the current frame while $h_f$ denotes the height.
\item Calculate the label $[O_1,O_2,O_3,O_4]$ for this sample as :
\begin{equation}
    \left\{
    \begin{array}{cc}
         &O_1 = \frac{\hat{x_1} - \frac{\hat{w_f}}{2}}{\hat{w_f}} = \frac{\hat{x_1}}{\hat{w_f}} - \frac{1}{2}  \\
         &O_2 = \frac{\hat{y_1} - \frac{\hat{y_f}}{2}}{\hat{h_f}} = \frac{\hat{y_1}}{\hat{h_f}} - \frac{1}{2}   \\
         &O_3 = \frac{\hat{x_2} - \frac{\hat{w_f}}{2}}{\hat{w_f}} = \frac{\hat{x_2}}{\hat{w_f}} - \frac{1}{2}  \\
         &O_4 = \frac{\hat{y_2} - \frac{\hat{y_f}}{2}}{\hat{h_f}} = \frac{\hat{y_2}}{\hat{h_f}} - \frac{1}{2}
    \end{array}
    \right.
\end{equation}

\noindent where $\hat{x_1}, \hat{y_1}, \hat{x_2}, \hat{y_2}$ denote the x-axis coordinate, y-axis coordinate of the top-left corner and the bottom-right corner of the target bounding box in the cropped image, which is the detection patch, respectively. $\hat{w_f}$ denotes the width of the detection patch while $\hat{h_f}$ denotes the height.
\end{enumerate}

Once we generate enough samples, we can start training the model. We use 239$\times$239, 125$\times$125 for the shape of two branch inputs. We set $w_x$ 15 and $w_z$ 7. Training the model within 5 epoches with the learning rate of 1e-3 and batch size of 80. The loss function we choose is Smooth L1 loss, which is:
\begin{equation}
    \label{con:eq3}
    Smooth_{l1}(x,\sigma)=
    \left\{
                 \begin{array}{lr}
                 0.5{\sigma}^2{x}^2, |x| \le \frac{1}{\sigma^2}&  \\
                 |x| - \frac{1}{2\sigma^2},  |x| \geq \frac{1}{\sigma^2}&  
                 \end{array}
    \right.
\end{equation}
According to Eq. \eqref{con:eq3}, the final loss is:
    \begin{equation}
        loss = \sum_{i=1}^{n} Smooth_{L1}(O_i^\star-O_i,\sigma)
    \end{equation}
\noindent where $O_i^\star$ denotes the groundtruth of the target position. 

\subsection{Inferring Phrase}
In the inferring phrase, we take the template patch and the detection patch as input, the output of the model is $[O_1,O_2,O_3,O_4]$ mentioned above. Then, we calculate the position according to the output and relevant frames. The detailed steps are as follows:
\begin{enumerate}
\item Crop the target in the first frame as the template patch, the template patch is fixed till the tracking progress ends. Record the position of template patch as [$left^\star, top^\star, width^\star, height^\star$].
\item Crop area according to ${left, top, right, bottom}$ of the current frame as detection patch by: 
\begin{equation}
\label{con:cropDetection}
    \left\{
    \begin{array}{cc}
         &left = max(0,left^\star-width^\star*\delta)  \\
         &top = max(0,top^\star-height^\star*\delta)  \\
         &right = max(left^\star+width^\star*(1+\delta),w_f)  \\
         &bottom = max(top^\star+height^\star*(1+\delta),h_f)  
    \end{array}
    \right.
\end{equation} \label{step_2}
\item Take the detection patch and the template patch into the tracker, and obtain the target region $x_1, y_1, w_1, h_1$ in the detection patch.
\item Considering the position of detection patch in the current frame, update the output as:
    \begin{equation}
        \left\{
        \begin{array}{cc}
            &x_1 = x_1 + left  \\
            &y_1 = y_1 + top \\
            &x_2 = x_2 + left  \\
            &y_2 = y_2 + top 
        \end{array}
        \right.
    \end{equation}
    
\item Update the position in current frame as [$left^\star, top^\star, width^\star, height^\star$].
\item go back to step \ref{step_2} till the tracking ends.
\end{enumerate}

\section{Experiment}
\label{sec:experiment}

In our experiment, we only use accuracy, robustness, EAO and EFO for our indicators in VOT2015 and VOT2016. In VOT2017, we run the real-time experiment and a running speed test for some excellent trackers. We add the speed test because the real-time experience show no difference when the speed of the trackers reach a certain value. For the comparative experimental results of other excellent trackers, we take the values for accuracy from the VOT reports to avoid the differences because of the implementation details. 

As for the OTB100, we draw the success plot and precision plot comparing with some other trackers, including some state-of-the-art trackers in these years and some fast traditional trackers.

Finally, we compare the size of our network with some other siamese tracking networks in the last sebsection.

\subsection{Result on VOT2015}
The VOT2015 provides fully-annotated sequences for visual object tracking. These sequences cover various scenarios, most of which correspond to one or more difficult problems in tracking. We test our tracker on the VOT2015 based on the rules, other tracker results for comparison are taken from the VOT2015 report. \\

\begin{table}[h!]
\normalsize
\centering
\caption{Comparsion with top trackers in VOT2015} 
\begin{tabular}{c c c c c}

Tracker& Accuracy& Robustness& EAO& EFO\\
\hline
DeepSRDCF& \bf{0.56}& \bf{1.0}& \bf{0.32}& 0.38\\
EBT& 0.47& 1.02& 0.31& 1.76\\
SRDCF& 0.56& 1.24& 0.29& 1.99\\
LDP& 0.51& 1.84& 0.28& 4.36\\
sPST& 0.55& 1.48& 0.28& 1.01\\
SC-EBT& 0.55& 1.86& 0.25& 0.80\\
\hline
Ours& 0.55& 1.95& 0.20& \bf{52.37}\\
\end{tabular}
\label{tab:2}

\end{table}

As shown in the Table \ref{tab:2}, we learn that most of trackers run slowly. Comparing to other trackers, our tracker achieves similar accuracy but works far faster.

\subsection{Result on VOT2016}
The VOT2016 shares the same sequences with VOT2015, just re-annotates the bounding boxes in the VOT2015 dataset. But there appears more excellent trackers in this year. The scores of trackers in this year are high in general, which may be caused by the similarity of the testing sequences. We select some excellent tracker for comparison. The indicators are the same as last subsection.

\begin{table}[h!]
\normalsize
\caption{Comparsion with top trackers in VOT2016} 
\centering
\begin{tabular}{c c c c c}
Tracker& Accuracy& Robustness& EAO& EFO\\
\hline
C-COT& 0.54& 0.24& \bf{0.33}& 0.51\\
TCNN& 0.56& 0.27& 0.33& 1.05\\
SSAT& \bf{0.58}& 0.29& 0.32& 0.48\\
MLDF& 0.49& \bf{0.23}& 0.31& 1.48\\
Staple& 0.54& 0.38& 0.30& 11.11\\
\hline
Ours& 0.51& 0.53& 0.19& \bf{53.05}\\
\end{tabular}
\label{tab:3}

\end{table}
As shown in Table \ref{tab:3}, in the sequences of VOT2016, C-COT performs best in our test. Our tracker runs faster than all of these trackers with good accuracy.

\subsection{Result on VOT2017}

EFO (Equivalent Filter Operations) reduces the variability by dividing the operation time of filtering on an image with fixed shape, which is used for indicating the running speed from VOT2014 to VOT2016. But the filtering operation could not represent all the operations in the different platform and different programming languages, which means EFO could not eliminate the variability. This is the reason why VOT do not use EFO after VOT2016. Instead, they propose a new experiment, real-time experiment, for testing other indicators with time limit. If the tracker cannot give the result of current frame in a given time, the testing system will use the last reported bounding box for the result. It means the evaluation result drops sharply if the running speed lower than the specified value.

But this experiment just divide the tracker into two parts, one part is slower than the specified value and the other part is faster. The differences between two trackers in one part can not be told. Therefore, we add an addition experiment for running some state-of-the-art trackers of VOT2017 on one machine.

\begin{figure}[ht]
\centering
\includegraphics[scale=0.5]{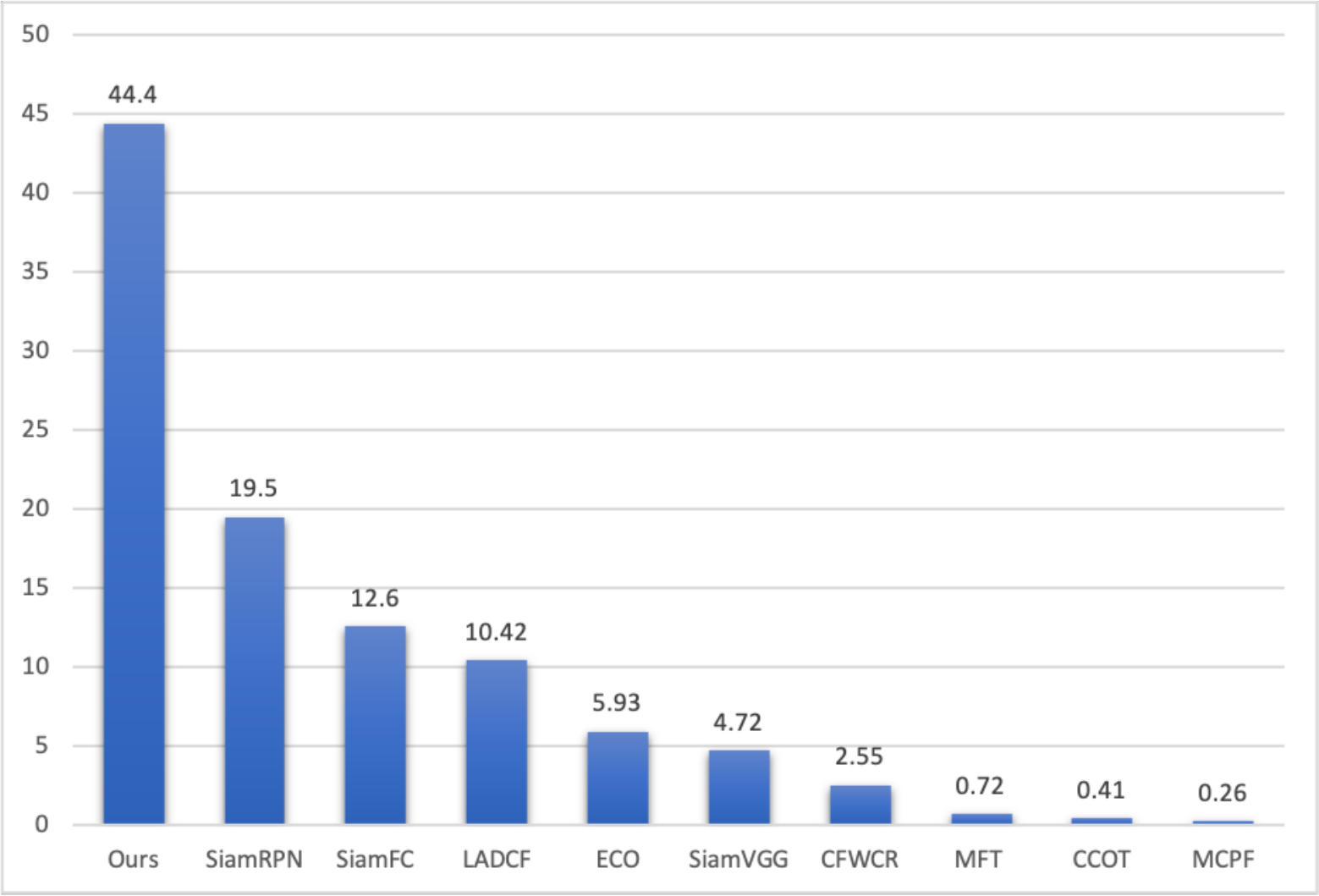}
\caption{Speed comparison of excellent trackers in VOT2017}
\label{fig_speed}
\end{figure}

The result of the real-time experiment of VOT2017 is shown in  Table \ref{tab:4}. In order to better demonstrate the speed advantage of our tracker, we make the test on a low-end device, the results are shown in Fig. \ref{fig_speed} and the unit is fps(frame per second). The configuration of the device is shown as Table \ref{tab:1}.
\begin{table}[h!]
\normalsize
\centering
\caption{Comparison with top trackers in VOT2017 real-time experiment} 
\begin{tabular}{c c c c}
Tracker& Accuracy& Robustness& EAO\\
\hline
CSRDCF++& 0.46& \bf{0.40}& \bf{0.21}\\
SiamFC& 0.50& 0.60& 0.18\\
ECOhc& 0.49& 0.57& 0.18\\
Staple& \bf{0.53}& 0.69& 0.17\\
ASMS& 0.49& 0.63& 0.17\\
\hline
Ours& 0.45& 0.81& 0.13\\
\end{tabular}
\label{tab:4}

\end{table}

\begin{table}[h!]
\normalsize
\centering
\caption{Hardware platform for experiments} 
\begin{tabular}{l l l l}
\centering

GPU& Processor\\
\hline
GeForce 840M& i7-4510U CPU @ 2.00GHz\\
\end{tabular}
\label{tab:1}
\end{table}
In VOT2017, some sequences in the data seems not challenging enough for state-of-the-art trackers, and hence are replaced by new sequences. From the reported results of the trackers, it is observable that sequences in VOT2017 are harder to track. But our tracker keeps the good performance. From the speed test, even on low-end device, our tracker still has an excellent speed, which is much faster than other trackers.

\subsection{Result on OTB100}
OTB2013 have two version, OTB50 and OTB100. OTB100 contains all of the sequences in OTB50. We use OTB100 to test our tracker for wider coverage. OTB100 evaluates trackers with one pass, which has not reinitialization after failure. This kind of evaluation is a big challenge for light network. In fact, this method is easy to waste the test set, because once the tracking fails, the following frames will be in vain.
\begin{center}
\begin{figure}[ht]
\centering
\includegraphics[scale=0.25]{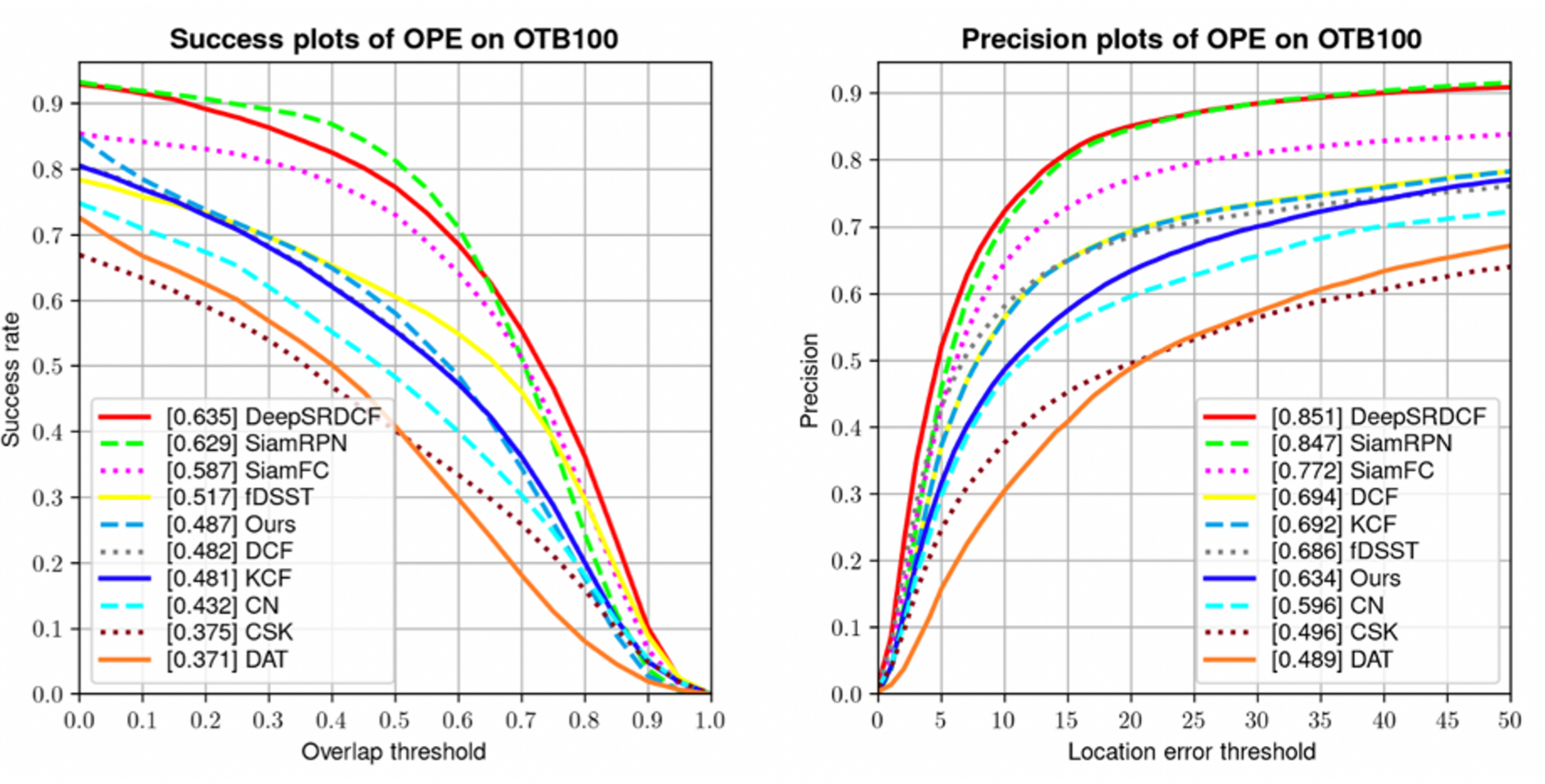}
\caption{Success plot and precision plot of OTB100}
\label{fig_otb}
\end{figure}
\end{center}

In this experiment, we compare with some excellent tracker in these years, such as DeepSRDCF, SiamRPN, SiamFC, fDSST. Aparting from this, we add some traditional trackers that are not state-of-art but light and fast for deploying and running just like ours. As shown in Fig. \ref{fig_otb}, even if OTB100 evaluates trackers with one pass, which has not reinitialization after failure, our tracker still have a good performance in success plot and precision plot.
\subsection{Model size}
Benefiting from the simplicity of our net, the model size is small enough. It makes it possible to deploy the model to low-end devices and even the mobile devices. We compare our model size with some networks based on siamese network. All of these models are end-to-end networks, the model size reflects the parameter size. The result is shown in Fig. 7.

\begin{figure}[ht]
    \centering
    \includegraphics[scale=0.5]{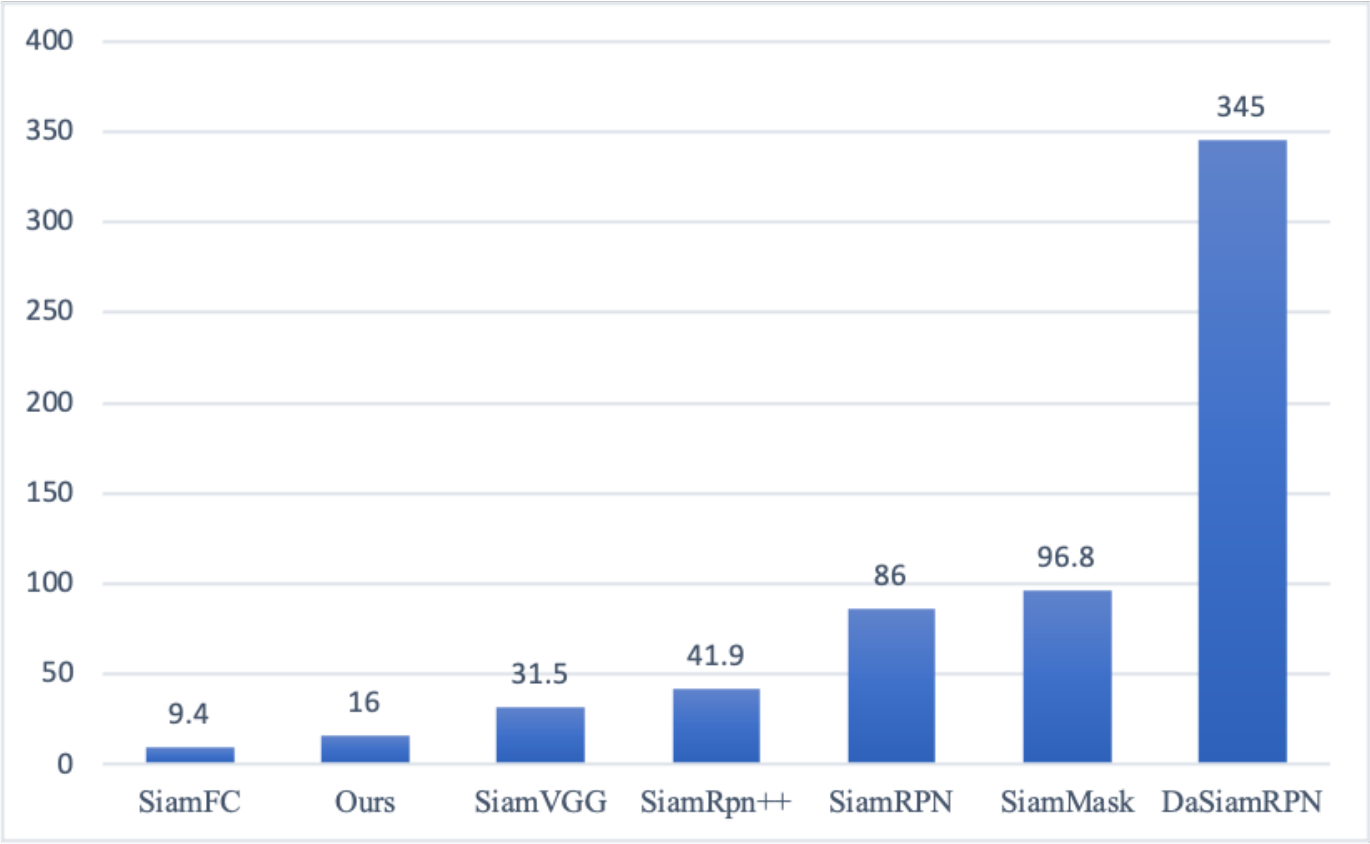}
    \caption{Size comparison of Siamese related models(Mb)}
    \label{fig_size}
\end{figure}
    
    As shown in Fig. \ref{fig_size}, our model is slighter than most of the siamese models. SiamFC has the lightest model, but the running speed is far lower than ours.

\section{Conclusion}
\label{sec:conclusion}

In this work, we propose a fast and effective object tracker based on siamese network. Our tracker is generated from an end-to-end network training on ImageNet-VID and GOT datasets. The network learned the positions for the top-left and right-bottom corners of the target. Our tracker could achieve real-time tracking even on low-end hardware, at the same time, it keeps the excellent accuracy comparing with other top trackers.

\bibliographystyle{abbrv}
\bibliography{main}

\end{document}